\documentclass[letterpaper, 10 pt, journal, dvipsnames, twoside]{IEEEtran}

\IEEEoverridecommandlockouts

\usepackage[utf8]{inputenc}
\usepackage{graphicx} \usepackage{amsmath} \usepackage{amssymb}  \usepackage{pgf}
\usepackage{url}
\usepackage{xcolor}
\usepackage{xspace}
\usepackage{enumitem}
\usepackage{multirow}
\usepackage{array}
\usepackage{tabularx}
\usepackage{textcomp}
\usepackage{mathtools}
\usepackage{bm}
\usepackage{adjustbox}
\usepackage{bbold}
\usepackage{footmisc}
\usepackage{microtype}

\usepackage{cite}

\makeatletter
\let\NAT@parse\undefined
\makeatother

\usepackage{hyperref}
\usepackage[capitalise]{cleveref}

\hypersetup{
  linkcolor  = NavyBlue,
  citecolor  = NavyBlue,
  urlcolor   = NavyBlue,
  colorlinks = true,
}

\newcolumntype{C}[1]{>{\centering\arraybackslash}m{#1}}
\newcolumntype{Y}{>{\centering\arraybackslash}X}

\graphicspath{{figs/}}

\makeatletter
\DeclareRobustCommand\onedot{\futurelet\@let@token\@onedot}
\def\@onedot{\ifx\@let@token.\else.\null\fi\xspace}

\def\eg{\emph{e.g}\onedot} 
\def\ie{\emph{i.e}\onedot}

\def\etal{\emph{et al}\onedot}
\makeatother

\title{How Many Events Do You Need?\\Event-based Visual Place Recognition\\Using Sparse But Varying Pixels}

\author{Tobias Fischer and Michael Milford\thanks{Manuscript received: July 7, 2022; Revised September 20, 2022; Accepted October 12, 2022.}
\thanks{This paper was recommended for publication by Editor S.~Behnke upon evaluation of the Associate Editor and Reviewers' comments. This work received funding from the Australian Government, via grant AUSMURIB000001 associated with ONR MURI grant N00014-19-1-2571, Intel via grant RV3.248.Fischer, and by funding from ARC Laureate Fellowship FL210100156 to MM. The authors acknowledge continued support from the Queensland University of Technology (QUT) through the Centre for Robotics.}\thanks{The authors are with the QUT Centre for Robotics, Queensland University of Technology, Brisbane, QLD 4000, Australia (e-mail: {\footnotesize \href{mailto:tobias.fischer@qut.edu.au}{tobias.fischer@qut.edu.au}, \href{mailto:michael.milford@qut.edu.au}{michael.milford@qut.edu.au}}).}\thanks{Digital Object Identifier (DOI): see top of this page.}}

\begin{document}
\bstctlcite{MyBSTcontrol}

\maketitle

\begin{abstract}

Event cameras continue to attract interest due to desirable characteristics such as high dynamic range, low latency, virtually no motion blur, and high energy efficiency. One of the potential applications that would benefit from these characteristics lies in visual place recognition for robot localization, i.e.~matching a query observation to the corresponding reference place in the database. In this letter, we explore the \emph{distinctiveness of event streams} from a small subset of pixels (in the tens or hundreds). We demonstrate that the absolute difference in the number of events at those pixel locations accumulated into event frames can be sufficient for the place recognition task, when pixels that display large variations in the reference set are used. Using such \emph{sparse} (over image coordinates) but \emph{varying} (variance over the number of events per pixel location) pixels enables frequent and computationally cheap updates of the location estimates. Furthermore, when event frames contain a constant number of events, our method takes full advantage of the event-driven nature of the sensory stream and displays promising robustness to changes in velocity. We evaluate our proposed approach on the Brisbane-Event-VPR dataset in an outdoor driving scenario, as well as the newly contributed indoor QCR-Event-VPR dataset that was captured with a DAVIS346 camera mounted on a mobile robotic platform. Our results show that our approach achieves competitive performance when compared to several baseline methods on those datasets, and is particularly well suited for compute- and energy-constrained platforms such as interplanetary rovers.


\end{abstract}
 \section{Introduction}
\label{sec:introduction}

\begin{figure}[t]
  \centering
  \includegraphics[width=0.98\linewidth]{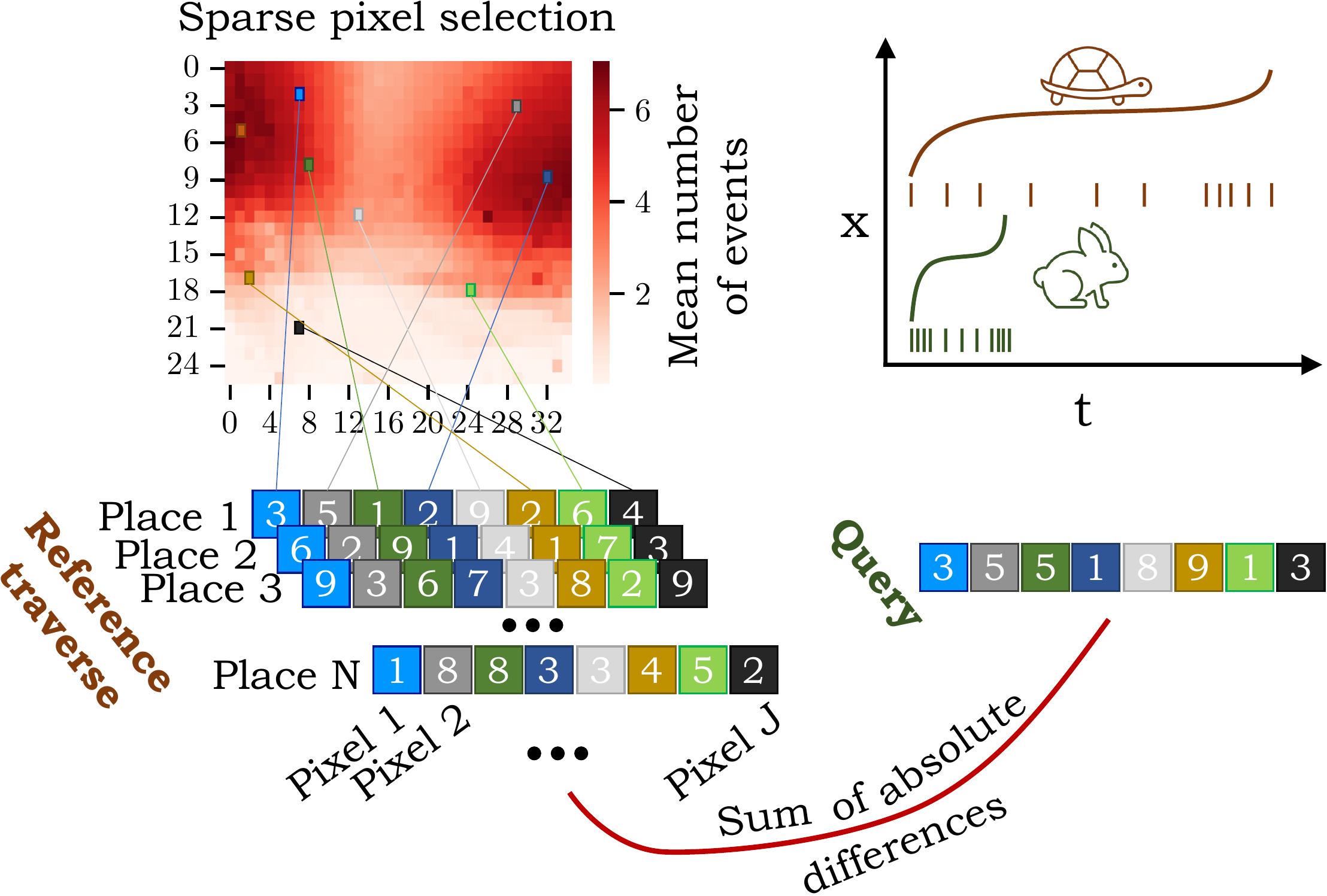}
  \vspace*{-0.25cm}
  \caption{\textbf{Event-based Place Recognition with Sparse Pixels.} Out of all available pixels, we select only a small subset (e.g.~0.028\%) to describe a place -- those that, on average, vary the most (top left). Each place descriptor contains the number of events that occurred at those sparse pixel locations (bottom left). Descriptor comparison is performed using the sum of absolute differences of these event counts (bottom right). If each feature descriptor contains a fixed number of events (summed over all sparse pixels), the feature descriptors demonstrate robustness to changes in velocity. This is due to the operating principle of event cameras, which captures the \emph{changes} in the scene regardless of how quickly they occur -- in an ideal scenario, the changes observed when going from place A to B are the same regardless of the velocity.}\label{fig:catchy_fig}\vspace*{-0.3cm}
\end{figure}

\IEEEPARstart{V}{isual} place recognition (VPR) is a crucial capability for autonomously navigating robots and vehicles~\cite{Garg2021,Masone2021,lowry2015visual}. In a nutshell, VPR is the task of retrieving the best matching reference observation out of a reference database, given a new observation at query time (Fig.~\ref{fig:catchy_fig}). If the score associated to that match is sufficiently high, the match can then be used to \eg~propose a loop closure within a Simultaneous Localization and Mapping (SLAM) system~\cite{cadena2016past}, or the VPR system can be used on its own to build a purely topological map~\cite{xu2021probabilistic}.

Our recent research has demonstrated the use of event cameras in the VPR task~\cite{Fischer2020}: their low power consumption and low bandwidth requirements make them well suited for on-board applications in robotic systems, and the high dynamic range and virtual absence of motion blur can be beneficial in low light conditions or high exposure to sunlight. At the same time, event cameras require new algorithms that can process the unconventional, and potentially very noisy, output data. They also have relatively low spatial resolution and are still comparatively expensive~\cite{Gallego2019}.

Each pixel of an event camera samples \textit{independently} and is equi-spaced in \textit{range} (\ie~an event is triggered when the change in the log intensity of a particular pixel exceeds a threshold), as opposed to conventional cameras where \textit{all} pixels are sampled equi-spaced in \textit{time}~\cite{Gallego2019}. This property implies that the event count of an ideal event camera is proportional to the dynamic changes that occur in the scene -- a static camera in a static scene would not output any events, while the total number of events is not impacted by how quickly the dynamic change occurs. Our first research question that we seek to answer is whether the dynamic change observed by traversing through a place remains relatively constant when the place is repeatedly traversed, potentially under changing appearance and velocities. 

As single events carry minimal localization-relevant information in isolation, various feature extraction techniques to accumulate events across space and time have been proposed in the literature, so that these features can be consumed in downstream tasks~\cite{rebecq2017evo,maqueda2018event,HFirst,HOTS,HATS,FEAST,gehrig2019end,TORE}. An alternative approach is to use recurrent neural networks to reconstruct conventional images from the event stream, so that conventional computer vision techniques can be applied -- often outperforming algorithms that operate ``directly'' on the event stream~\cite{e2vid,Scheerlinck2020,pan2020high}. However, many of these feature extractors are workload or memory intensive, which offsets some of the computational and energy advantages of event cameras. Our second research question is thus whether we can describe a place in a storage and computationally friendly manner, while still achieving high performance and being robust to appearance variations and image misalignments.

Our third research question asks whether we can use just a few pixels of event camera footage to obtain such a high performing system, and whether we can find \emph{discriminative} image regions (\eg~the side of the road), opposed to less informative regions like the dashboard of a car.

\vspace{0.1cm}
\noindent Our contributions can be summarized as follows:
\vspace{-0.05cm}
\begin{enumerate}
    \item We introduce a novel high-performing, computationally lightweight, training-free event-based pipeline for visual place recognition that is robust to moderate velocity changes (see Fig.~\ref{fig:catchy_fig}). The pipeline operates on a small subset of the stream of events, specifically on pixels that exhibit a high variation in the event counts during mapping time, combined with ``surround suppression'' to avoid selection of neighboring pixels, which exhibit a high correlation and thus add little information. \item We compare our method to several baselines~\cite{Fischer2020,Milford_2015_CVPR_Workshops} and provide extensive ablation studies that evaluate the impact of the number of pixels that are selected, the location of those pixels in the image space, the time duration of the event frames, and the sequence length.
    \item We collect the QCR-Event-VPR dataset, which contains 16 traverses of a 160m long route traveled by a Clearpath Jackal robot with a DAVIS346 camera mounted \mbox{forward-,} side-, or downward-facing.
\end{enumerate}

\noindent To foster future research, we make the code and QCR-Event-VPR dataset available: \url{https://github.com/Tobias-Fischer/sparse-event-vpr}. \section{Related Works}
\label{sec:relatedworks}
We review works on representing events in \mbox{Section~\ref{subsec:rw_eventrepresentations}} before providing an overview of related works on visual place recognition in Section~\ref{subsec:rw_vpr}, and detailing prior works on place recognition using event cameras in Section~\ref{subsec:rw_eventvpr}.

\subsection{Event Representations}
\label{subsec:rw_eventrepresentations}
A range of feature representations have been proposed for event data, many of them focusing on the object classification task. Event frames~\cite{rebecq2017evo} are the simplest representation, where a conventional single-channel image is formed, with each pixel containing the number of events that occurred at that location over either a fixed time period, a fixed number of events across the image, or using adaptive times slices with variable duration~\cite{Liu2018}. If positive and negative polarities are considered separately, a two-channel image is formed~\cite{maqueda2018event}.

Lagorce \etal~\cite{HOTS} developed the Hierarchy Of Time Surfaces (HOTS) that describe the recent activity within a local neighboring window, whereby less recent events exponentially decay in value. Sironi \etal~proposed Histograms of Averaged Time Surfaces (HATS)~\cite{HATS} which is less sensitive to noise and contrast when compared to HOTS. FEAST~\cite{FEAST} is an extension of HOTS with adaptive thresholds, which allows for a simpler network architecture and integrated noise filtering.

Gehrig \etal~\cite{gehrig2019end} presented a unifying framework that learns the event representation end-to-end in a deep network. Most recently Baldwin \etal~\cite{TORE} introduced Time-Ordered Recent Event (TORE) volumes to compactly store raw spike timing information. Detailed comparisons of different event representations are provided in~\cite{TORE,gu2022mdoe}; and \cite{jiao2021comparing} compares event representations in the SLAM context.

To the best of our knowledge, we are not aware of any other works in the event camera literature that make use of just a tiny fraction of the available pixels. The closest related works in that regard are~\cite{cannici2019attention,renner2019event} which focus on estimating the region that an object encompasses to recognize~\cite{cannici2019attention} and track~\cite{renner2019event} the object using attention mechanisms. In addition, a variety of works explore using as few events as possible to e.g.~update event-based graph networks~\cite{li2021graph,schaefer2022aegnn} or spiking neural networks~\cite{gehrig2020event} \emph{in the temporal domain}, but in contrast to our work, still use all available pixels \emph{spatially}. Note that the sparse event representation in our work is task-specific and not likely to be applicable in other tasks without modifications.

\subsection{Visual Place Recognition}
\label{subsec:rw_vpr}
Visual place recognition (VPR) is a thriving area, with applications including autonomous driving, robot navigation and augmented reality~\cite{Garg2021,Masone2021,lowry2015visual}. One particular application of VPR is in the proposal of loop closure candidates within Simultaneous Localization and Mapping (SLAM)~\cite{cadena2016past}, where VPR is used to recognize previously visited locations to build a globally accurate map. Challenges for VPR include changes in the appearance of the scene due to the time of day, season, and weather; structural changes due to occlusions and new constructions; and viewpoint changes~\cite{lowry2015visual}. In this article, we focus mainly on appearance changes, which are predominant in road-based scenarios where both translation and rotation changes are limited~\cite{Garg2021}.

A reoccurring theme in VPR research is the use of sequences to improve performance. The underlying motivation is that perceptual aliasing -- the phenomenon that two distinct places look nearly identical -- is less likely to occur when considering multiple observations over time. In the SeqSLAM~\cite{milford2012seqslam}, a sequence score is obtained as the sum of the difference scores within a time window, and two sequences are matched if the lowest difference score is significantly lower than the second lowest difference score (ratio test). A large body of work has since focused on sequence matching~\cite{Milford_2015_CVPR_Workshops,han2018sequence,siam2017fast,hansen2014visual}, and most recently sequence-based methods have been adapted to better work with deep-learned features~\cite{garg2022seqmatchnet}.

The most closely related work in the conventional VPR literature is \cite{milford2013vision} that investigates the amount of visual information that is required for place recognition in conventional cameras by varying the number of pixels, their pixel depth, and the sequence length. They were able to successfully localize in a path-like office environment with just two pixels measuring light intensity, however requiring a relatively large sequence length of 200 frames in that case. Our paper goes beyond~\cite{milford2013vision} by using sparse pixels, rather than down-sampling image regions, and explores unique properties of event cameras, which are not covered in~\cite{milford2013vision}.

\subsection{Event-based Visual Place Recognition}
\label{subsec:rw_eventvpr}
Before introducing our proposed approach, we will review the related works in the VPR domain that use event cameras. In the initial approach by Milford \etal~\cite{Milford2015}, temporally binned windows were formed into event frames~\cite{rebecq2017evo}, which were then patch-normalized and compared using sequence matching. Rather than using the whole image, here we show that the informativeness of certain areas in the image is non-uniformly distributed.

In our prior work~\cite{Fischer2020}, we first reconstructed conventional frames from the event stream via~\cite{e2vid,Scheerlinck2020}, extracted features from these frames using NetVLAD~\cite{Arandjelovic2018}, and then found the best match by comparing the query and reference features using the cosine distance. The key novelty was in using an ensemble of features which emerged by using varying temporal windows to reconstruct the frames. 

Kong \etal~\cite{Kong2022} took a similar approach named Event-VPR where the NetVLAD layer is trained directly on a voxel grid representation~\cite{gehrig2019end}. Another recent interesting approach is EventVLAD~\cite{lee2021eventvlad} which estimates edge images from the event stream and subsequently trains the NetVLAD layer using these edge images. Rather than going the roundabout way via reconstructing conventional frames or applying computationally expensive deep learning methods, in this research we directly use the event stream to recognize places.

Works in related areas that could inform event-based visual place recognition include \cite{mueggler2018continuous,liu2022async}, which tackle event-based visual odometry, and \cite{chamorro2022event,kim2016real}, which introduced event-based SLAM algorithms, however lack a loop closure component.  \section{Proposed Approach}
\label{sec:methods}

\subsection{Preliminaries and Notation}
\label{subsec:notation}
We closely follow the notation that was used in our prior work~\cite{Fischer2020}. The $i$-th event is denoted as $\mathbf{e}_i = (\mathbf{u}_i,t_i,p_i)$, where $\mathbf{u}_i=(u_i,v_i)^T$ refers to the pixel location where the event occurred, $t_i$ is the timestamp associated to the event, and $p_i\in\{-1,1\}$ is the event's polarity that indicates positive or negative brightness change.
As reviewed in Section~\ref{subsec:rw_eventrepresentations}, prior work~\cite{rebecq2017evo} formed event frames $\mathbf{I}$ by accumulating events over a specified time window, so that:
\begin{equation}\label{eq:event_frames}
\mathbf{I}_{u,v}(t_k) := \big\{ \vert \mathbf{e}_i \vert \bigm| u_i = u \wedge v_i = v \wedge t_i \in [t_k, t_k + \tau) \big\},
\end{equation}
where $t_{k+1} = t_k + \tau$.  Alternatively, events can also be accumulated so that each event frame contains a fixed number of events $N$ (and event frames $k$ and $k+1$ are non-overlapping): 
\begin{equation}\label{eq:event_frames_fixed}
\sum_{u,v} \mathbf{I}_{u,v}(k) = N.
\end{equation}
In both cases, by considering the polarity $p$, two channels $\mathbf{I}^+$ (containing only events with $p=1$) and $\mathbf{I}^-$ ($p=-1$) can be created~\cite{maqueda2018event}. Unless otherwise stated, in the remainder of this paper, we accumulate events over a specified time window as in Eq.~(\ref{eq:event_frames}) and do not consider the polarity $p$.

\subsection{Sparse Pixel Selection}
\label{subsec:pixelselection}
\begin{figure}[t]
  \centering
  \includegraphics[width=0.9\linewidth]{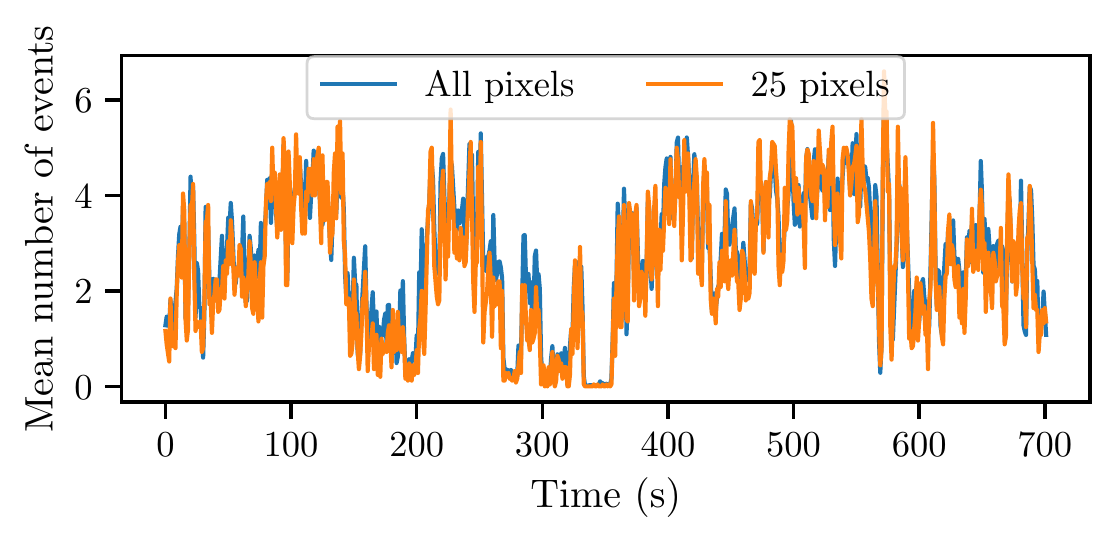}
  \vspace*{-0.3cm}
  \caption{\textbf{Event profile is maintained when drastically sub-sampling pixels.} This figure shows the number of events that occurred within a one second window, averaged either across all pixels (blue line) or when considering just 0.028\% of the pixels (25 out of 89960 overall; orange line), on the Brisbane-Event-VPR dataset. One can observe a close correlation, indicating that the subset is representative of all pixels.}\vspace*{-0.2cm}
  \label{fig:mean_events_per_second}\end{figure}

In this work, rather than considering all pixels $\mathbf{u} \in \left(\{1,\ \cdots, W\} \times \{1,\ \cdots, H\}\right)$, where $W$ and $H$ indicate the width and height of the image respectively, we instead select a small subset $\mathbf{S} \subsetneq \mathbf{I}$ of pixels. The motivation behind this choice is that we observed that small subsets $\mathbf{S}$ are representative of the whole image, as illustrated in Fig.~\ref{fig:mean_events_per_second}. We define $S_{u,v}$ as the variance in the number of events observed at pixel location $(u,v)$ in the reference traverse\footnote{We note that using simply the mean number of events results in very similar behavior and performance, as mean and variance in this particular case are highly correlated.}:
\begin{equation}
    S_{u,v} = \frac{1}{K} \sum_{k=1}^K \big(\mathbf{I}_{u,v}(t_k) - \mu\big)^2,
    \label{eq:variance}
\end{equation}
where $\mu$ is the mean number of events and $K$ is the number of event frames.

To select the subset of pixels, we draw a sequence of samples based on the following probability mass function $p(u, v)$:
\begin{gather}
        p(u, v) = P_{XY}(X=u, Y=v) = \frac{S_{u,v}}{A}, \text{where}
\end{gather}
$A=\sum_{u=1}^{W}\,\sum_{v=1}^{H}S_{u,v}$ is a normalization constant.

One could now sample from this probability mass function without replacement. However, to avoid selection of neighboring pixels, whose event rate is highly correlated, we employ surround suppression modeled by the inverse of a Gaussian probability density function. The methodology is detailed in the remainder of this subsection; and code implementing the sparse pixel selection is provided in the associated repository. 

\begin{figure}[t]
  \centering
  \includegraphics[width=0.62\linewidth]{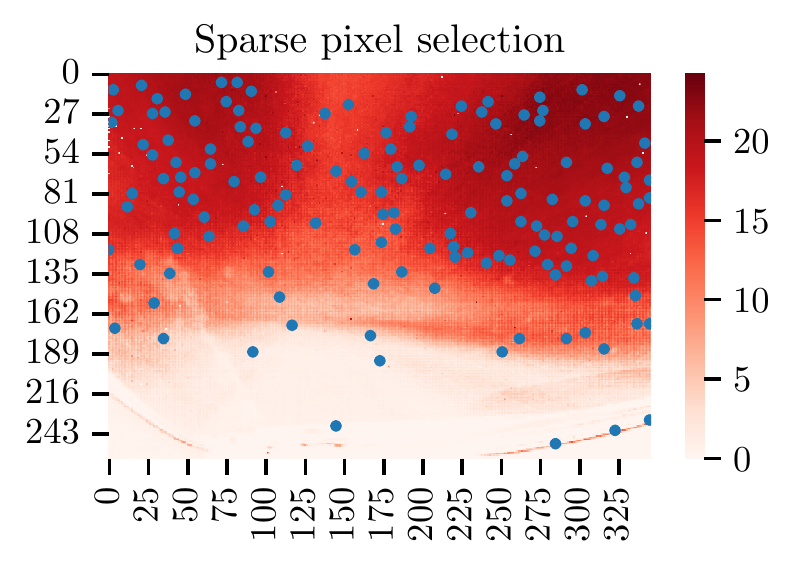}
  \vspace*{-0.1cm}
  \caption{\textbf{Event occurrences are unevenly distributed, and pixels with higher variance are selected.} To create this figure, we first accumulated events in $\tau=1s$ windows across the whole ``Sunset 1'' reference traverse in the Brisbane-Event-VPR dataset into event windows. This figure shows the variance in the number of events per pixel across these event windows (see color bar). An example of the subset of pixels that were selected according to the probabilistic scheme described in Section~\ref{subsec:pixelselection} are shown with blue dots. The dark red area indicates that most selected pixels are in regions with a high variance (darker red indicates higher variance), while they are also spatially separated due to the surround inhibition. The light red areas at the bottom and center of the event frame correspond to the bonnet and street surface respectively, which (on average) have a much lower variance than the areas on the side where substantial vegetation and buildings are present.}\vspace*{-0.2cm}
  \label{fig:salient_pixel_selection}\vspace*{-0.2cm}
\end{figure}

Specifically, we first define the discrete bivariate Gaussian
\begin{gather}
    \widetilde{T}_{XY}(u, v, \sigma) = T_X(u, \sigma)\ T_Y(v, \sigma),\label{eq:bivariate_gaussian}
\end{gather}
where 
$T_X(u, \sigma)$ and $T_Y(v, \sigma)$ are independent univariate Gaussian kernels along the $u$ and $v$ coordinates, respectively. The standard deviation $\sigma$ of the Gaussian kernels is chosen to be the same across the $u$ and $v$ dimensions\footnote{Note that the variance $S_{u,v}$ in Eq.~(\ref{eq:variance}) and $\sigma$ in Eq.~(\ref{eq:bivariate_gaussian}) are not related. $S_{u,v}$ describes how much the event count varies over time, while $\sigma$ describes the width of the Gaussian kernels that are used for surround suppression.}. The joint probability mass function $\hat{p}_{j+1}(u_{j+1}, v_{j+1})$ to select the $(j+1)$-th pixel in a recursive manner is then:
\begin{gather}
    \hat{p}_{j+1}(u_{j+1}, v_{j+1}) = \widetilde{A}_{j+1}\cdot\widetilde{G}_{XY}(u_j, v_j, \sigma)\cdot\hat{p}_{j}(u_j, v_j),
\end{gather}
where $\widetilde{G}_{XY}$ is the inverse distribution of $\widetilde{T}_{XY}$, $(u_j, v_j)$ denotes the coordinates of the $j$-th selected pixel, $\hat{p}_{0}(u_0, v_0) = p(u, v)$, and $\widetilde{A}_{j+1}$ is a normalization constant so that $\hat{p}_{j+1}(u_{j+1}, v_{j+1})$ is a valid probability mass function.
We repeat this recursive process until we find a subset of pixels $\mathbf{S}$, where $\mathbf{card(S)}=J$. Fig.~\ref{fig:salient_pixel_selection} shows an example of the selected sparse pixels.

\subsection{Place Recognition Using a Small Subset of Pixels}
We follow the typical place recognition setup where $\mathbf{S}^r(t_k)$ denotes the number of events observed at the selected pixels $\mathbf{S}$ at the $k$-th place in the reference traverse (we use superscripts $r$ and $q$ to indicate reference and query traverses respectively). The aim is that given $\mathbf{S}^q(t_j)$ at query time, we want to find the best matching place $k$ in $\mathbf{S}^r$.

The data association is performed using the sum-of-absolute-differences, as widely used for conventional camera images~\cite{hansen2014visual,hausler2019multi,milford2012seqslam}, but comparing the number of events rather than the intensity value, and only considering the subset of pixels:
\begin{gather}
    \label{eq:sad}
    \mathbf{D}(j,k) = \sum_{(u,v)\in\mathbf{S}}\big\lvert\mathbf{S}^r(t_k)-\mathbf{S}^q(t_j)\big\rvert,
\end{gather}
where $\mathbf{D}(j,k)$ refers to the sum-of-absolute-differences between reference location $j$ and query location $k$.

\subsection{Using Spatio-Temporal Sequences}
\label{subsec:sequences}
As the observed number of events within a few selected pixels is relatively low, and thus the expressiveness of a single difference value is limited, we rely on sequence matching~\cite{milford2012seqslam} to combine the temporal information contained in $L$ query event frames. Specifically, we use the computationally efficient 2D convolution-based implementation proposed in~\cite{garg2022seqmatchnet}:
\begin{align}
    \mathbf{D}_{seq}(j,k) = \mathbb{1}_L * \mathbf{D}_{seq}(j,k) / L,
\end{align}
where $L$ describes the sequence length and $\mathbb{1}_L$ denotes the identity matrix of dimension $L\times L$. 

In the experimental results, we show an ablation study that investigates the trade-off between using event frames that accumulate over longer time windows $\tau$ (or equivalently event frames with a bigger number of fixed events $N$) and using sequence matching that accumulates individual event frames over an equivalent time.
 \section{Experimental Setup}
\label{sec:setup}

In this section, we first describe the datasets that we use in our experiments, namely the newly collected QCR-Event-VPR dataset and our previously introduced Brisbane-Event-VPR dataset (Section~\ref{subsec:datasets}). We then set out the evaluation metrics in Section~\ref{subsec:evaluation}, and describe the baseline methods and pre-processing steps in Sections~\ref{subsec:comparison} and~\ref{subsec:preprocessing} respectively. Finally, we present implementation details in Section~\ref{ref:implementation}.

\subsection{Datasets}
\label{subsec:datasets}
We previously introduced the Brisbane-Event-VPR dataset~\cite{Fischer2020} which was recorded in the suburb of Brookfield in Brisbane, Australia. The recording setup was such that a DAVIS346 camera was mounted forward-facing behind the windshield of a Honda Civic. The same 8km route was traversed six times at different times of the day and under different weather conditions. Following~\cite{Fischer2020}, we do not evaluate on the night-time traverse, as the camera parameters were not adjusted for these difficult lighting conditions. We extend Brisbane-Event-VPR to use the provided GPS information, instead of the manual labeling that was used in~\cite{Fischer2020}.

\begin{figure}[t]
  \vspace*{0.1cm}
  \centering
  \includegraphics[width=0.7\linewidth]{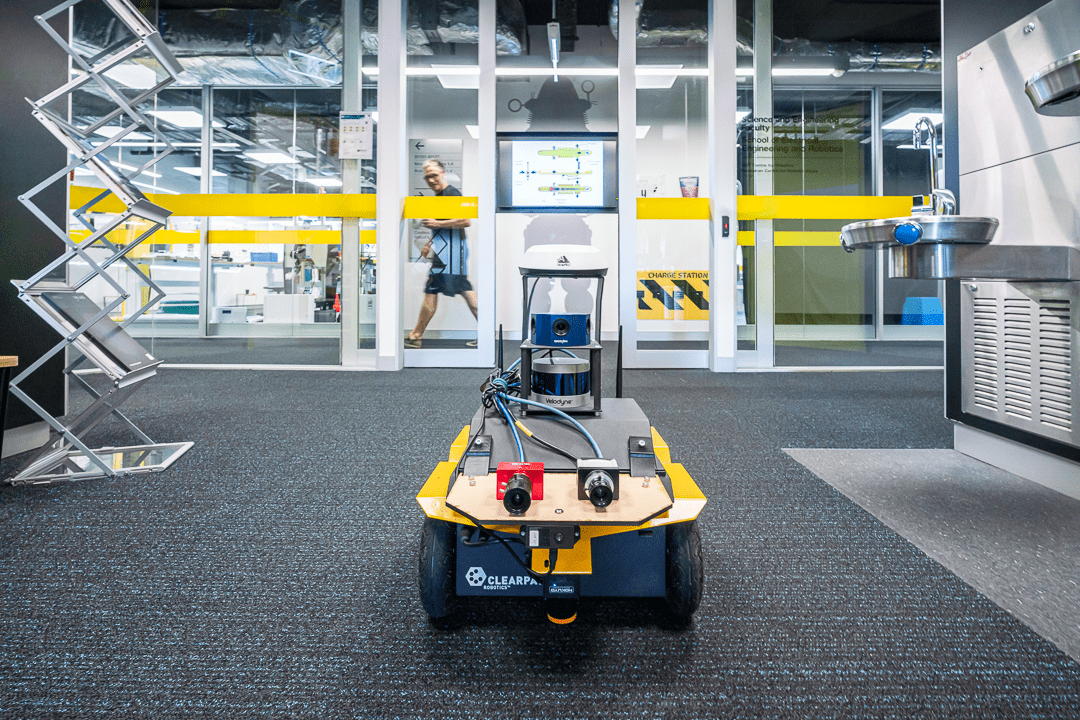}
  \caption{\textbf{Recording setup for the QCR-Event-VPR dataset.} A DAVIS346 event camera (in red) was mounted on a Clearpath Jackal mobile robot. The camera can be installed forward-, side-, or downward-facing.}\label{fig:QCREventVPR}\vspace*{-0.2cm}
\end{figure}

In this letter, we contribute the QCR-Event-VPR dataset, which was captured indoors in the QUT Centre for Robotics. The DAVIS346 was mounted on a Clearpath Jackal robot using a flexible mount that enabled us to have the camera forward-, side-, or downward-facing. The same 160m route was traversed 16 times at different speeds and times of the day. We collected and evaluate on this dataset to demonstrate that our method generalizes to different recording setups, as well as to easily and safely adjust the velocities in which the route was traversed in (ranging from 154s to 262s, i.e.~70\% difference in velocities). Fig.~\ref{fig:QCREventVPR} shows the recording setup.

\subsection{Evaluation Metrics}
\label{subsec:evaluation}
We use the standard information retrieval-based metrics that are commonly used in place recognition~\cite{Garg2021,Fischer2020}: precision $P=tp/(tp+fp)$ and recall $R=tp/(tp+fn)$. To create precision-recall curves, we sweep over the distance values in Eq.~\ref{eq:sad} for the system to decide whether to accept the match. 

We report the precision at 100\% recall (P@100R, i.e.~what is the percentage of correct matches when the system is forced to match all query images to a database image) and recall at 99\% precision (i.e.~at the virtual absence of false data associations, how many query images were matched to database images). Following~\cite{Fischer2020,Kong2022}, we allow a tolerance of 70 meters around the ground truth location as correct matches on the Brisbane-Event-VPR dataset. For the QCR-Event-VPR dataset, the tolerance is 3 meters.

\subsection{Baseline Methods}
\label{subsec:comparison}
We provide comparisons to several baseline methods: Firstly, the sum-of-absolute-differences of event counts when using all pixels (similar to \cite{Milford2015} but without using patch normalization). For fair comparison, we use the same temporal binning as in our method (Eqs.~\ref{eq:event_frames} and~\ref{eq:event_frames_fixed}). Secondly, we compare to two versions of our previous method, Ensemble-Event-VPR~\cite{Fischer2020}, that relies on the reconstruction of conventional images~\cite{scheerlinck2020fast} and extracting NetVLAD features~\cite{Arandjelovic2018} from these reconstructed images. In one version, we use an ensemble that combines temporal window sizes of varying length, and in another we use a more computationally efficient method that uses a single temporal window size of 30ms to reconstruct the images. We note that the reconstructed images are of lower spatial resolution (346x260 pixels) and quality (see~\cite{Fischer2020}) compared to conventional images taken with e.g.~a smartphone camera, which impacts NetVLAD's performance. Finally, we compare to Event-VPR by Kong \etal~\cite{Kong2022}\footnote{We note that Event-VPR~\cite{Kong2022} is unfairly disadvantaged as it does not use sequences. However, \cite{Kong2022} has shown that it performs similar to Ensemble-Event-VPR~\cite{Fischer2020}, and thus the performance of \cite{Fischer2020} (where sequence matching is used for fair comparisons) can be used as a proxy for \cite{Kong2022}.\label{footnote:kongsequence}} and EventVLAD by Lee and Kim~\cite{lee2021eventvlad} (see Section~\ref{subsec:rw_eventvpr} for a description of these methods).

\subsection{Pre-processing}
\label{subsec:preprocessing}
We follow the same pipeline as in~\cite{Fischer2020} to remove bursts that are erroneously triggered for the majority of pixels within a very short time when sunlight hits the DAVIS346 bias generator, and to remove hot pixels. Such processing is needed due to the relative immaturity of event cameras, and is common in the event camera literature~\cite{scheerlinck2018continuous,Fischer2020}.

\subsection{Implementation}
\label{ref:implementation}
We use Python alongside the NumPy, PyTorch and Tonic~\cite{lenz_gregor_2021_5079802} libraries to implement our method. Unless otherwise noted, we use a time window length of $\tau=1s$. We select $J=150$ pixels based on Fig.~\ref{fig:performance_number_of_pixels} and the equivalent experiments for the other reference-query traverse pairs, which show that this is a high performing region. We use a sequence length of $L=5$. The results are averaged over 5 trials; the selected pixels differ in each of the trials according to the random sampling strategy introduced in Section~\ref{subsec:pixelselection}. All experiments are conducted on a MacBook Pro M1 without using a GPU\footnote{We do not make use of a GPU as our aim is a generically applicable method that can run on robotic systems that are power constrained.}. \section{Results}
\label{sec:results}

In this section, we present results that answer our three research questions that we set out in Section~\ref{sec:introduction}. Specifically, we first demonstrate that an intelligent selection of sparse pixels as introduced in Section~\ref{subsec:pixelselection} outperforms a random pixel selection (Section~\ref{subsec:comparestrategies}). We then demonstrate that our approach is robust to moderate pixel image misalignments (Section~\ref{subsec:robustnessviewpoint}) and velocity variations (Sections~\ref{subsec:robustnessvelocity}). We also compare the performance of our approach with two baseline methods, SAD and NetVLAD, and show that our approach performs similarly in terms of recall performance (Section~\ref{subsec:comparisonsota}) but being computationally more lightweight (Section~\ref{subsec:computational_efficiency}). Finally, we briefly discuss whether our method is also applicable to conventional images (Section~\ref{subsec:conventionalcomparison}).

\subsection{Comparing Selection Strategies}
\label{subsec:comparestrategies}
Here we set out to compare our selection of highly varying pixels (Section~\ref{subsec:pixelselection}) with a random selection of pixels. We repeat each experiment five times, and report results on the Brisbane-Event-VPR dataset as the traverses are significantly longer than in the QCR-Event-VPR dataset. We select a sequence length of 5 (noting that we observed similar performance differences for other sequence lengths that are not shown due to space constraints).

\begin{figure}[t]
  \centering
  \includegraphics[width=0.85\linewidth]{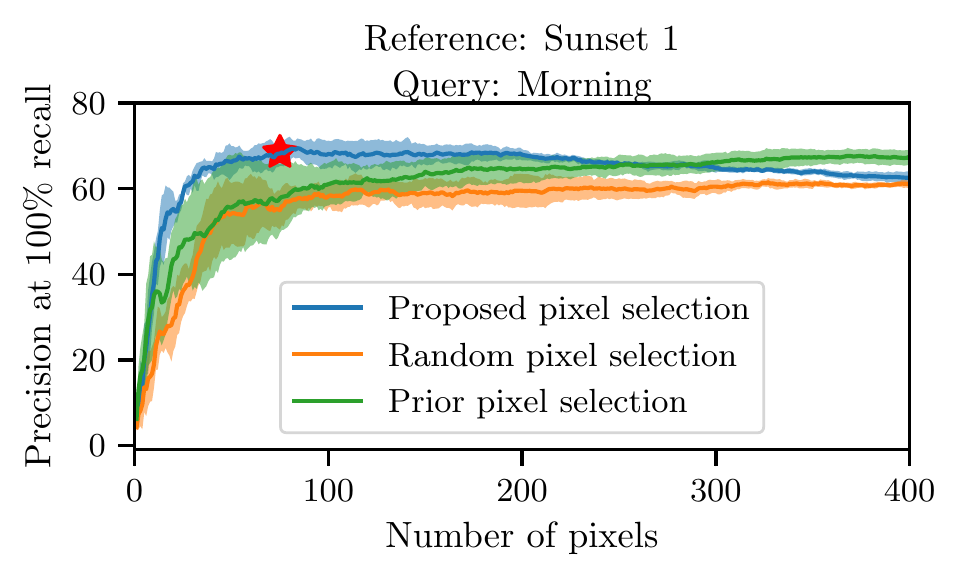}
  \vspace*{-0.2cm}
  \caption{\textbf{Comparing random with highly varying, discriminative pixels.} This figure simultaneously investigates the performance increase with an increasing number of pixels that are used to describe a place, and whether highly varying pixels (blue) perform better than random pixels (orange) or discarding the bottom third of the event frames a priori (green). The red star indicates the highest P@100R performance overall.}\label{fig:performance_number_of_pixels}\vspace*{-0.2cm}
\end{figure}

Figure~\ref{fig:performance_number_of_pixels} provides an example comparison on one reference (sunset 1) query (morning) combination. There are three interesting findings. Firstly, selecting highly varying pixels generally outperforms the random pixel selection, as well as a random pixel selection where with a priori knowledge no pixels are selected in the bottom third of the event frames (see Fig.~\ref{fig:salient_pixel_selection}). This is more pronounced the fewer pixels are being used. Secondly, adding more and more pixels does not necessarily result in a performance increase -- in fact, in this particular example the best performance (highlighted by a red star) is obtained using just 83 pixels. Thirdly, combining the first two findings, the performance advantage of using highly varying pixels vanishes when using a large number of pixels. Averaged over the four different query traverses, the R@99P increased on average by 11.1\%, and the P@100R by 5.3\%.

\subsection{Robustness to Pixel Image Misalignment}
\label{subsec:robustnessviewpoint}
In our approach, we essentially compare the number of events that have occurred at particular pixel locations between the query and reference traverses. But what if the camera location changes between the traverses, so that pixel ($u, v$) at reference time does not correspond to the exact same pixel location at query time? We investigate this by artificially shifting all pixels ($u, v$) in the query traverse by $(\Delta u, \Delta v)$ so that $u^*=u+\Delta u$ and $v^*=v+\Delta v$, while leaving the reference traverse untouched.

Before presenting the results of this investigation, we want to note that 1) implicitly the robustness is shown in the Brisbane-Event-VPR results, as the camera mount was slightly moved between each recording session, and 2) the resulting pixel image misalignments are relatively minor, as opposed to major shifts in viewpoint which are still an open research problem in the visual place recognition field~\cite{Garg2021}.

\begin{figure}[t]
  \centering
  \includegraphics[height=0.53\linewidth]{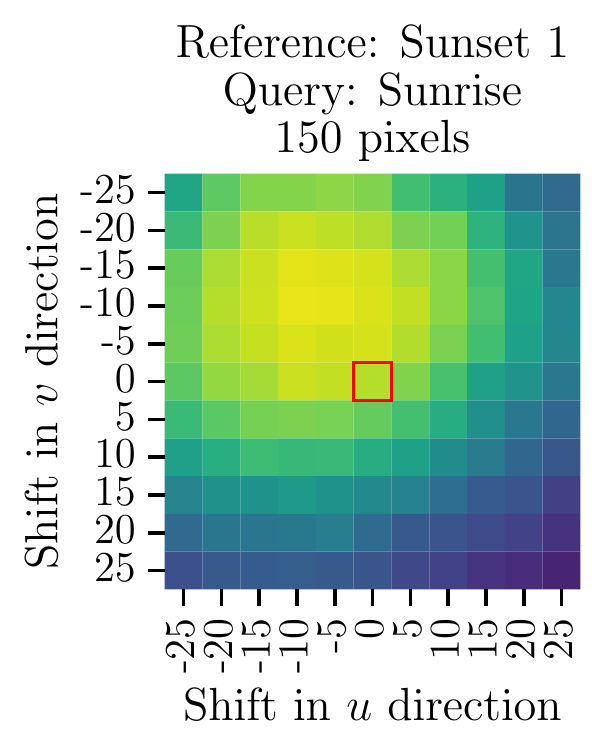}
  \includegraphics[height=0.53\linewidth]{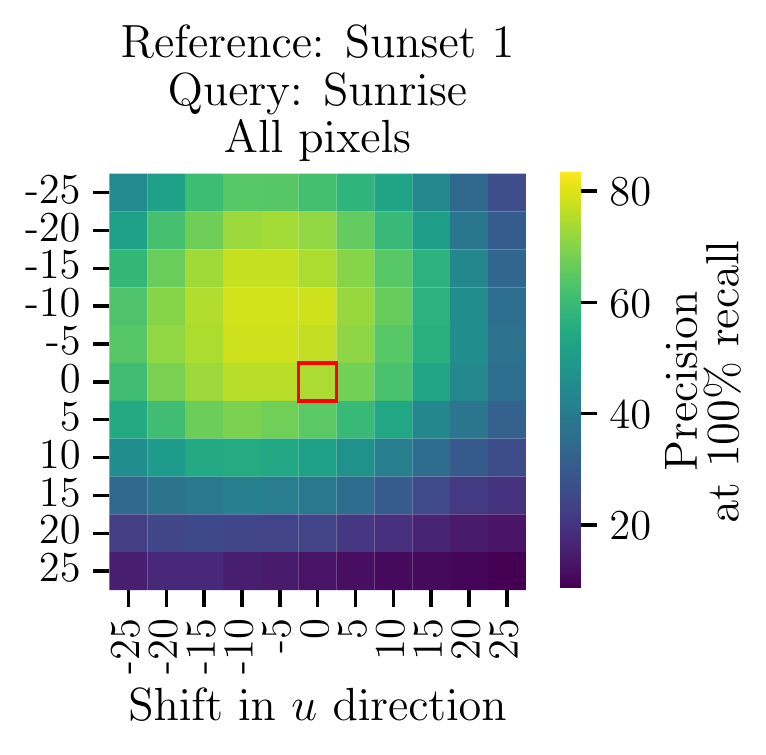}
  \vspace*{-0.2cm}
  \caption{\textbf{Robustness to pixel image misalignments.} This figure shows that the performance (as measured by P@100R) of our method (left plot) gracefully decays as the pixel shift increases. The performance decay is comparable to that of standard sum-of-absolute-differences where all pixels are used (right plot). Interestingly, the performance decay is not symmetric around the center (shown with a red border); the highest performance is observed at pixel offset $(-10, -15)$, indicating that there was a shift in camera position between reference and query times.}\label{fig:shift_pixels}\vspace*{-0.1cm}
\end{figure}

The results for this investigation is summarized in Fig.~\ref{fig:shift_pixels}. There are two key take-aways: Firstly, the performance degrades gracefully, and secondly, the robustness to pixel image misalignments of our sparse method that uses just 150 pixels is roughly the same as when using all 89960 pixels.

\begin{figure}[t]
  \centering
  \includegraphics[width=0.99\linewidth]{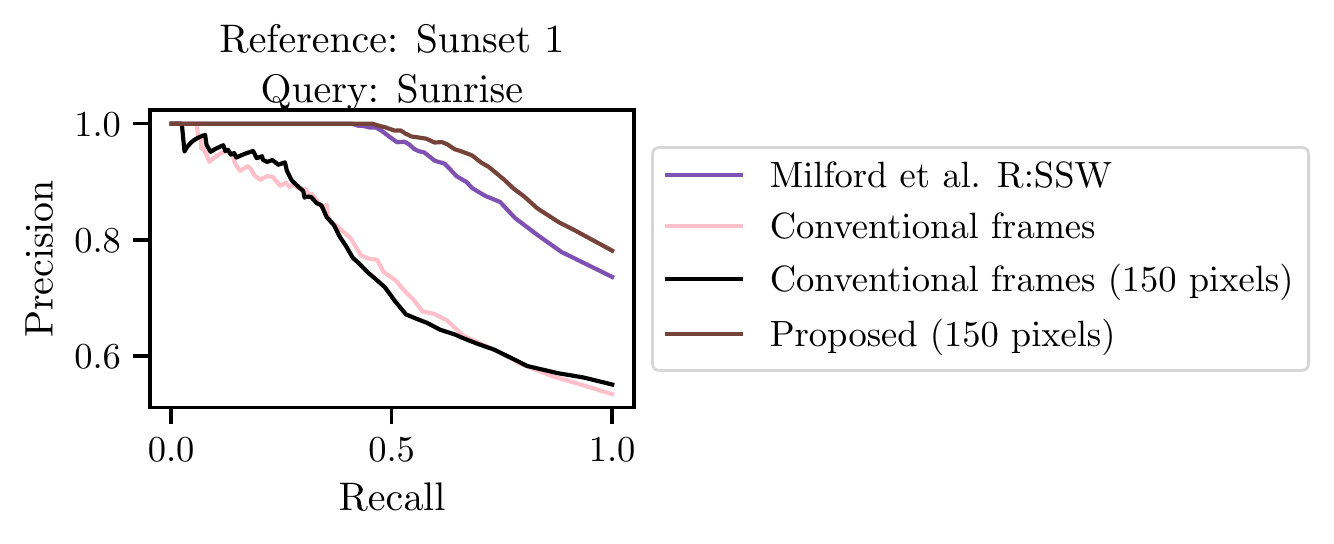}
\caption{\textbf{Application of sparse pixel selection to conventional images.} This figure investigates the performance of sparse pixel selection on conventional images. Similar performance trends as in the event-based case can be observed, with all pixels (pink) performing similarly to the sparse pixel selection (black). The performance on conventional images is significantly lower compared to the event stream, however noting that the DAVIS346 native RGB frames are of low quality.}\label{fig:conventional}\end{figure}

\begin{figure*}[t]
\includegraphics[height=0.18\linewidth]{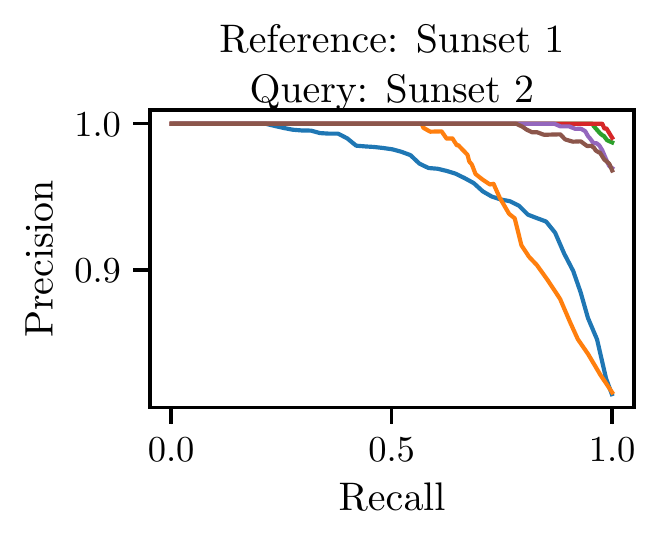}
  \hspace{0.21cm}
  \includegraphics[height=0.18\linewidth]{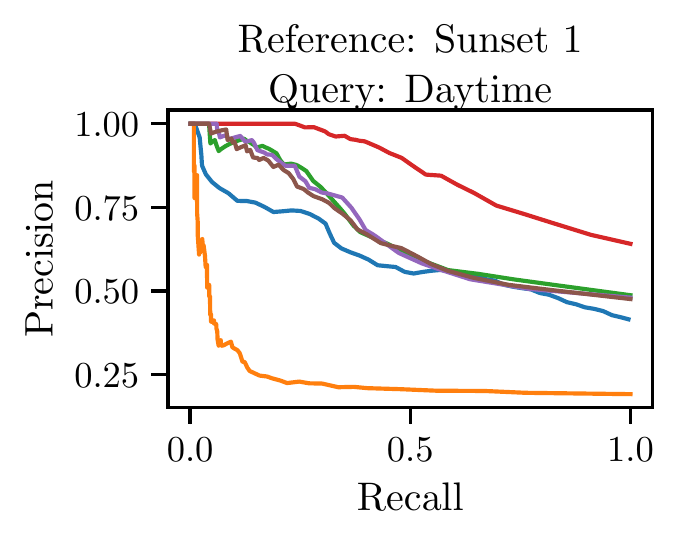}
  \hspace{0.21cm}
  \includegraphics[height=0.18\linewidth]{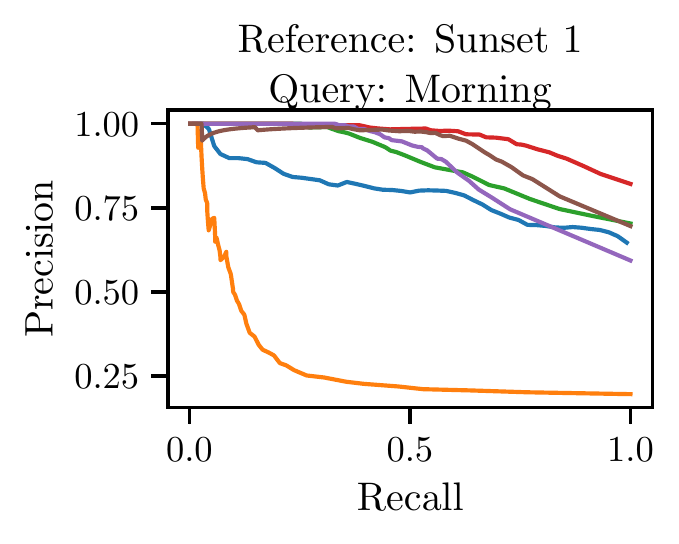}
  \hspace{0.21cm}
  \includegraphics[height=0.18\linewidth]{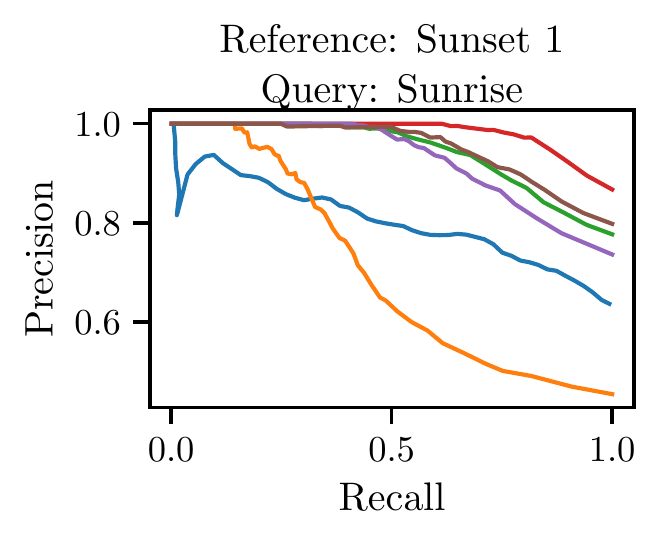}
\\[-0.1cm]
\includegraphics[height=0.18\linewidth]{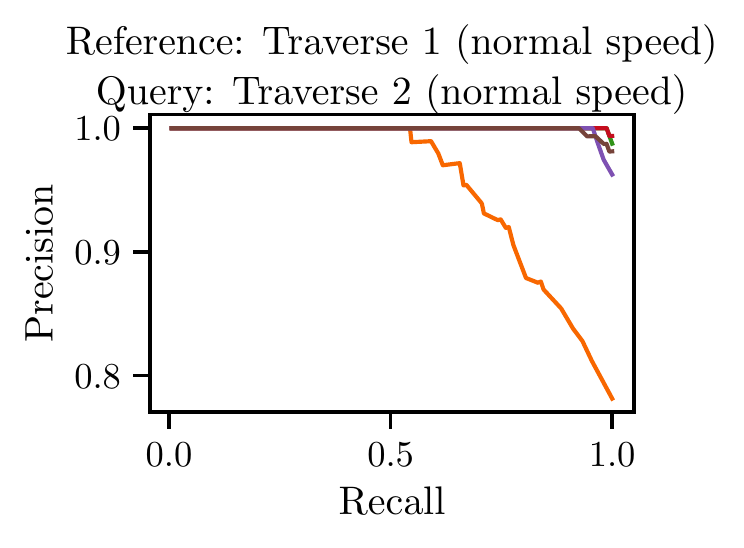}
\includegraphics[height=0.18\linewidth]{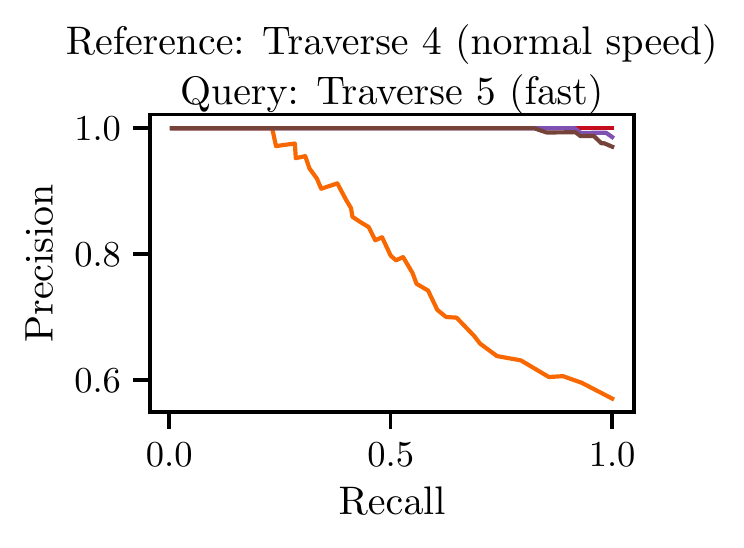}
\includegraphics[height=0.18\linewidth]{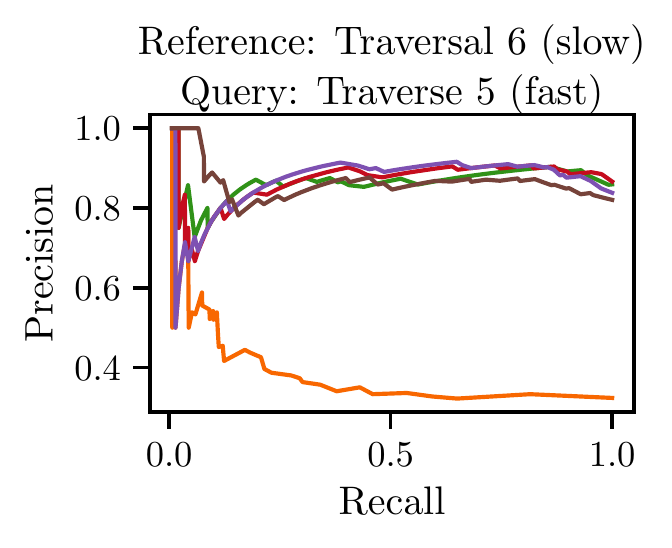}
\raisebox{0.6cm}{\includegraphics[trim={1mm 1mm 1mm 1mm},clip,width=0.275\linewidth]{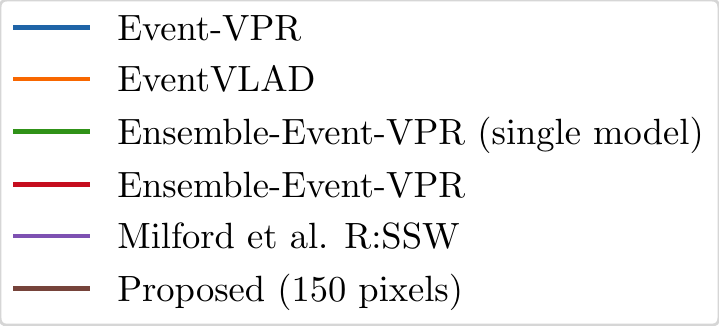}}
\vspace*{-0.5cm}
\caption{\textbf{Precision-recall curves.} Top row: Results for the Brisbane-Event-VPR dataset. Following~\cite{Fischer2020}, the Sunset 1 traverse was used as reference traverse, and the remaining four traverses as query traverses. Our proposed method using 150 selected pixels (brown) generally outperforms using all pixels~\cite{Milford2015} (purple) and EventVLAD~\cite{lee2021eventvlad} (orange). Our method performs roughly on par with the much more computationally expensive Ensemble-Event-VPR~\cite{Fischer2020} when one of their models is used individually (green), but is outperformed by their even more expensive ensemble variant~\cite{Fischer2020} (red). Bottom row: Results on the QCR-Event-VPR dataset. The general trend of these experiments follows those observed on the Brisbane-Event-VPR dataset: Our method performs competitively while being significantly faster at inference time. Specifically, the left plot shows a traverse pair where the robot traveled at the same speed, the middle plot shows moderate speed variation, and in the right plot there is a large variation in speeds between the query and reference traverses.}\label{fig:prcurve}\vspace*{-0.15cm}
\end{figure*}

\subsection{Robustness to Velocity Changes}
\label{subsec:robustnessvelocity}
In our first research question, we asked ourselves whether the amount of dynamic change remains relatively constant when a place is repeatedly traversed. As defined in Section~\ref{subsec:notation}, we accumulate events either over a given time or by splitting event frames such that each frame contains a fixed number of events $N$. It is often argued that the latter way preserves the data-driven nature of event cameras~\cite{gallego2017accurate}. Indeed, in traverses that have significant velocity differences within the QCR-Event-VPR dataset, one can observe that the total number of events across the whole traverse remains roughly the same ($\pm 10\%$), while the duration changes significantly ($70\%$ change between reference and query traverse). This also implies that the dynamic change within each event frame remains similar between the reference and query traverses despite the velocity differences. 

Indeed using event frames with a fixed $N$ results in much higher performance, i.e.~a P@100R of $79.0\pm1.4\%$ (averaged over 5 trials) compared to $20.6\pm1.3\%$ when using fixed $\tau$. For fair comparison, we 1) used a temporal window size that is as long as the effective average window size when keeping $N$ constant, 2) kept all other hyper-parameters the same, and 3) ensured that the same subset of pixels was used. In summary, our method displays promising robustness to change in velocity, which will be relevant in high-speed applications as further discussed in Section~\ref{sec:conclusions}.

\subsection{Comparison to State-of-the-Art}
\label{subsec:comparisonsota}
Precision-recall curves for comparisons with the state-of-the-art are shown in Fig.~\ref{fig:prcurve}. Generally speaking, our method outperforms SAD~\cite{Milford2015} and EventVLAD~\cite{lee2021eventvlad} on all traverses. It also outperforms Event-VPR~\cite{Kong2022}, however as noted in Footnote~\ref{footnote:kongsequence}, Event-VPR is disadvantaged as sequence matching is not used. Our proposed method shows very similar performance when compared to a single model of Ensemble-Event-VPR~\cite{Fischer2020}, i.e.~first reconstructing conventional images and applying NetVLAD on those images. The ensemble of 9 models as proposed in~\cite{Fischer2020} outperforms our method, but is not applicable in practical scenarios due to the high computational demands as discussed next. These observations hold on both the Brisbane-Event-VPR dataset and the QCR-Event-VPR dataset. 

\subsection{Computational Efficiency}
\label{subsec:computational_efficiency}
Our method using sparsely selected pixels is 50 times faster compared to SAD~\cite{Milford2015}, with the calculation of Eq.~\ref{eq:sad} taking ${\approx}1$ms compared to ${\approx}50$ms when using all pixels. The computational advantage is even more pronounced in the comparison to~\cite{Fischer2020}: ${\approx}1$ms for our method compared to over ${\approx}1000$ms for a single model in~\cite{Fischer2020} (image reconstruction: ${\approx}10$ms, NetVLAD feature extraction ${\approx}1000$ms, distance matrix calculation: ${\approx}8$ms) and over ${\approx}2000$ms when considering an ensemble of 9 models as proposed in~\cite{Fischer2020}. We note that both EventVLAD~\cite{lee2021eventvlad} and Event-VPR~\cite{Kong2022} have similar runtimes (i.e.~${\approx}1000$ms) as the single model version of~\cite{Fischer2020} due to their architectural similarity. In summary, our method achieves competitive performance and can easily be deployed on systems that do not have a GPU.

\subsection{Application to Conventional Images}
\label{subsec:conventionalcomparison}
In this section, we apply our sparse pixel selection to conventional camera-based place recognition. Instead of using the variance in the event counts, we simply used the variance in the intensity of the native RGB frames that are provided by the DAVIS346 camera (the DAVIS346 can simultaneously record events and RGB frames) in the selection process.

Fig.~\ref{fig:conventional} shows that using the native RGB frames of the DAVIS camera (pink line) led to lower performance compared to using the event-based approach (purple line). We also found that the sparse pixels can be used as place recognition descriptors, even in the case of RGB frames (black line). Indeed, just like in the event-based case (brown line), the pixel subsampling led to similar performance as using all pixels, while being computationally more efficient. We note that these findings match those in~\cite{Fischer2020} where a similar experiment that compared conventional image data with event data for visual place recognition was conducted. We want to emphasize that the native RGB frames provided by the DAVIS346 are of low quality, which affects the results.

 \section{Discussion and Conclusions}
\label{sec:conclusions}
In this paper, we presented a novel training-free method for event-based visual place recognition. Our main goal was to introduce a computationally lightweight method that is applicable in mobile robots. Despite our method's simplicity and not requiring any training, we have demonstrated competitive performance when compared to the state-of-the-art. To foster future research in this area, we make both the QCR-Event-VPR dataset and our code available to the community.

Throughout the paper we have answered our research questions: We found that the dynamic change when traversing through a place remains relatively constant even under moderate changes of appearance, velocity and viewpoint (research question 1). We have further demonstrated that the event count of a small subset of pixels is sufficient to succinctly describe a place -- which has the practical benefit of very low demands with respect to storage and compute (research question 2). Finally, we have shown that a high performing visual place recognition system can be obtained when using discriminative image regions -- i.e.~those that have a large variance in the number of events across the reference traverse (research question 3).

There are several directions for future works. Firstly, we could use adaptive duration event frames~\cite{Liu2018} or motion-corrected edge-like images via contrast maximization~\cite{gallego2018unifying}, as opposed to the simple accumulation of events to form event frames. It would also be interesting to use overlapping event frames at query time, which would enable place matching on an event-by-event basis. Secondly, our method could benefit from a more rigorous information-theory driven approach to the pixel selection~\cite{fontan2020information}. Thirdly, we are interested in taking the precise timing and order of the events into account~\cite{dauwels2008similarity}. Taken together with sophisticated odometry~\cite{xu2021probabilistic}, this could pave the way to place recognition in ultrasonar applications. Finally, we could imagine to apply our methodology to other camera types, including light-field cameras that can trade-off spatial and temporal resolution~\cite{gupta2010flexible,tambe2013towards} -- or even design cheap sensors that only capture a sparse representation of the scene.

{
\hbadness 10000\relax
\bibliographystyle{IEEEtran}
\bibliography{references}

\begin{thebibliography}{10}
\providecommand{\url}[1]{#1}
\csname url@samestyle\endcsname
\providecommand{\newblock}{\relax}
\providecommand{\bibinfo}[2]{#2}
\providecommand{\BIBentrySTDinterwordspacing}{\spaceskip=0pt\relax}
\providecommand{\BIBentryALTinterwordstretchfactor}{4}
\providecommand{\BIBentryALTinterwordspacing}{\spaceskip=\fontdimen2\font plus
\BIBentryALTinterwordstretchfactor\fontdimen3\font minus
  \fontdimen4\font\relax}
\providecommand{\BIBforeignlanguage}[2]{{%
\expandafter\ifx\csname l@#1\endcsname\relax
\typeout{** WARNING: IEEEtran.bst: No hyphenation pattern has been}%
\typeout{** loaded for the language `#1'. Using the pattern for}%
\typeout{** the default language instead.}%
\else
\language=\csname l@#1\endcsname
\fi
#2}}
\providecommand{\BIBdecl}{\relax}
\BIBdecl

\bibitem{Garg2021}
S.~Garg, T.~Fischer, and M.~Milford, ``{Where is your place, Visual Place
  Recognition?}'' in \emph{IJCAI}, 2021, pp. 4416--4425.

\bibitem{Masone2021}
C.~Masone and B.~Caputo, ``A survey on deep visual place recognition,''
  \emph{IEEE Access}, vol.~9, pp. 19\,516--19\,547, 2021.

\bibitem{lowry2015visual}
S.~Lowry \emph{et~al.}, ``Visual place recognition: A survey,'' \emph{IEEE
  Trans. Robot.}, vol.~32, no.~1, pp. 1--19, 2015.

\bibitem{cadena2016past}
C.~Cadena \emph{et~al.}, ``Past, present, and future of simultaneous
  localization and mapping: Toward the robust-perception age,'' \emph{IEEE
  Trans. Robot.}, vol.~32, no.~6, pp. 1309--1332, 2016.

\bibitem{xu2021probabilistic}
M.~Xu, T.~Fischer, N.~S{\"u}nderhauf, and M.~Milford, ``Probabilistic
  appearance-invariant topometric localization with new place awareness,''
  \emph{IEEE Robot. Autom. Lett.}, vol.~6, no.~4, pp. 6985--6992, 2021.

\bibitem{Fischer2020}
T.~Fischer and M.~Milford, ``{Event-Based Visual Place Recognition With
  Ensembles of Temporal Windows},'' \emph{IEEE Robot. Autom. Lett.}, vol.~5,
  no.~4, pp. 6924--6931, 2020.

\bibitem{Gallego2019}
G.~Gallego \emph{et~al.}, ``Event-based vision: A survey,'' \emph{IEEE Trans.
  Pattern Anal. Mach. Intell.}, vol.~44, no.~1, pp. 154--180, 2020.

\bibitem{rebecq2017evo}
H.~Rebecq, T.~Horstsch{\"a}fer, G.~Gallego, and D.~Scaramuzza, ``{EVO: A}
  geometric approach to event-based 6-dof parallel tracking and mapping in real
  time,'' \emph{IEEE Robot. Autom. Lett.}, vol.~2, no.~2, pp. 593--600, 2017.

\bibitem{maqueda2018event}
A.~I. Maqueda, A.~Loquercio, G.~Gallego, N.~Garc{\'\i}a, and D.~Scaramuzza,
  ``Event-based vision meets deep learning on steering prediction for
  self-driving cars,'' in \emph{IEEE Conf. Comput. Vis. Pattern Recog.}, 2018,
  pp. 5419--5427.

\bibitem{HFirst}
G.~Orchard \emph{et~al.}, ``{HFirst: A} temporal approach to object
  recognition,'' \emph{IEEE Trans. Pattern Anal. Mach. Intell.}, vol.~37,
  no.~10, pp. 2028--2040, 2015.

\bibitem{HOTS}
X.~Lagorce, G.~Orchard, F.~Galluppi, B.~E. Shi, and R.~B. Benosman, ``{HOTS: A
  Hierarchy Of event-based Time-Surfaces} for pattern recognition,'' \emph{IEEE
  Trans. Pattern Anal. Mach. Intell.}, vol.~39, no.~7, pp. 1346--1359, 2016.

\bibitem{HATS}
A.~Sironi, M.~Brambilla, N.~Bourdis, X.~Lagorce, and R.~Benosman, ``{HATS:
  Histograms of Averaged Time Surfaces} for robust event-based object
  classification,'' in \emph{IEEE Conf. Comput. Vis. Pattern Recog.}, 2018, pp.
  1731--1740.

\bibitem{FEAST}
S.~Afshar \emph{et~al.}, ``Event-based feature extraction using adaptive
  selection thresholds,'' \emph{Sensors}, vol.~20, no.~6, p. 1600, 2020.

\bibitem{gehrig2019end}
D.~Gehrig, A.~Loquercio, K.~G. Derpanis, and D.~Scaramuzza, ``End-to-end
  learning of representations for asynchronous event-based data,'' in
  \emph{IEEE Int. Conf. Comput. Vis.}, 2019, pp. 5633--5643.

\bibitem{TORE}
R.~Baldwin, R.~Liu, M.~M. Almatrafi, V.~K. Asari, and K.~Hirakawa,
  ``Time-ordered recent event {(TORE)} volumes for event cameras,'' \emph{IEEE
  Trans. Pattern Anal. Mach. Intell.}, 2022, to appear.

\bibitem{e2vid}
H.~Rebecq, R.~Ranftl, V.~Koltun, and D.~Scaramuzza, ``High speed and high
  dynamic range video with an event camera,'' \emph{IEEE Trans. Pattern Anal.
  Mach. Intell.}, vol.~43, no.~6, pp. 1964--1980, 2021.

\bibitem{Scheerlinck2020}
C.~Scheerlinck, H.~Rebecq, N.~Barnes, R.~E. Mahony, and D.~Scaramuzza, ``{Fast
  Image Reconstruction with an Event Camera},'' in \emph{IEEE Winter Conf.
  Appl. Comput. Vis.}, 2020, pp. 156--163.

\bibitem{pan2020high}
L.~Pan \emph{et~al.}, ``High frame rate video reconstruction based on an event
  camera,'' \emph{IEEE Trans. Pattern Anal. Mach. Intell.}, vol.~44, no.~5, pp.
  2519--2533, 2022.

\bibitem{Milford_2015_CVPR_Workshops}
M.~Milford \emph{et~al.}, ``Sequence searching with deep-learnt depth for
  condition- and viewpoint-invariant route-based place recognition,'' in
  \emph{IEEE Conf. Comput. Vis. Pattern Recog. Worksh.}, 2015.

\bibitem{Liu2018}
M.~Liu and T.~Delbruck, ``Adaptive time-slice block-matching optical flow
  algorithm for dynamic vision sensors,'' in \emph{Brit. Mach. Vis. Conf.},
  2018.

\bibitem{gu2022mdoe}
F.~Gu \emph{et~al.}, ``{MDOE: A} spatiotemporal event representation
  considering the magnitude and density of events,'' \emph{IEEE Robot. Autom.
  Lett.}, vol.~7, no.~3, pp. 7966--7973, 2022.

\bibitem{jiao2021comparing}
J.~Jiao \emph{et~al.}, ``Comparing representations in tracking for event
  camera-based {SLAM},'' in \emph{IEEE Conf. Comput. Vis. Pattern Recog.
  Worksh.}, 2021, pp. 1369--1376.

\bibitem{cannici2019attention}
M.~Cannici, M.~Ciccone, A.~Romanoni, and M.~Matteucci, ``Attention mechanisms
  for object recognition with event-based cameras,'' in \emph{IEEE Winter Conf.
  Appl. Comput. Vis.}, 2019, pp. 1127--1136.

\bibitem{renner2019event}
A.~Renner, M.~Evanusa, and Y.~Sandamirskaya, ``Event-based attention and
  tracking on neuromorphic hardware,'' in \emph{IEEE Conf. Comput. Vis. Pattern
  Recog. Worksh.}, 2019.

\bibitem{li2021graph}
Y.~Li \emph{et~al.}, ``Graph-based asynchronous event processing for rapid
  object recognition,'' in \emph{IEEE Int. Conf. Comput. Vis.}, 2021, pp.
  934--943.

\bibitem{schaefer2022aegnn}
S.~Schaefer, D.~Gehrig, and D.~Scaramuzza, ``{AEGNN: Asynchronous} event-based
  graph neural networks,'' in \emph{IEEE Conf. Comput. Vis. Pattern Recog.},
  2022, pp. 12\,371--12\,381.

\bibitem{gehrig2020event}
M.~Gehrig, S.~B. Shrestha, D.~Mouritzen, and D.~Scaramuzza, ``Event-based
  angular velocity regression with spiking networks,'' in \emph{IEEE Int. Conf.
  Robot. Autom.}, 2020, pp. 4195--4202.

\bibitem{milford2012seqslam}
M.~J. Milford and G.~F. Wyeth, ``{SeqSLAM}: Visual route-based navigation for
  sunny summer days and stormy winter nights,'' in \emph{IEEE Int. Conf. Robot.
  Autom.}, 2012, pp. 1643--1649.

\bibitem{han2018sequence}
F.~Han, H.~Wang, G.~Huang, and H.~Zhang, ``Sequence-based sparse optimization
  methods for long-term loop closure detection in visual {SLAM},''
  \emph{Autonomous Robots}, vol.~42, no.~7, pp. 1323--1335, 2018.

\bibitem{siam2017fast}
S.~M. Siam and H.~Zhang, ``Fast-{SeqSLAM}: A fast appearance based place
  recognition algorithm,'' in \emph{IEEE Int. Conf. Robot. Autom.}, 2017, pp.
  5702--5708.

\bibitem{hansen2014visual}
P.~Hansen and B.~Browning, ``Visual place recognition using {HMM} sequence
  matching,'' in \emph{IEEE/RSJ Int. Conf. Intell. Robot. Syst.}, 2014, pp.
  4549--4555.

\bibitem{garg2022seqmatchnet}
S.~Garg, M.~Vankadari, and M.~Milford, ``{SeqMatchNet}: Contrastive learning
  with sequence matching for place recognition \& relocalization,'' in
  \emph{Conf. Robot. Learn.}, 2022, pp. 429--443.

\bibitem{milford2013vision}
M.~Milford, ``Vision-based place recognition: how low can you go?'' \emph{Int.
  J. Robot. Res.}, vol.~32, no.~7, pp. 766--789, 2013.

\bibitem{Milford2015}
M.~Milford, H.~Kim, S.~Leutenegger, and A.~Davison, ``{Towards Visual SLAM with
  Event-based Cameras},'' in \emph{Robotics: Science and Systems Workshops},
  2015.

\bibitem{Arandjelovic2018}
R.~Arandjelovic, P.~Gronat, A.~Torii, T.~Pajdla, and J.~Sivic, ``{NetVLAD: CNN
  Architecture for Weakly Supervised Place Recognition},'' \emph{IEEE Trans.
  Pattern Anal. Mach. Intell.}, vol.~40, no.~6, pp. 1437--1451, 2018.

\bibitem{Kong2022}
D.~Kong \emph{et~al.}, ``Event-{VPR}: {End}-to-end weakly supervised deep
  network architecture for visual place recognition using event-based vision
  sensor,'' \emph{{IEEE} Trans. Instrum. Meas.}, vol.~71, pp. 1--18, 2022.

\bibitem{lee2021eventvlad}
A.~J. Lee and A.~Kim, ``{EventVLAD: Visual} place recognition with
  reconstructed edges from event cameras,'' in \emph{IEEE/RSJ Int. Conf.
  Intell. Robot. Syst.}, 2021, pp. 2247--2252.

\bibitem{mueggler2018continuous}
E.~Mueggler, G.~Gallego, H.~Rebecq, and D.~Scaramuzza, ``Continuous-time
  visual-inertial odometry for event cameras,'' \emph{IEEE Trans. Robot.},
  vol.~34, no.~6, pp. 1425--1440, 2018.

\bibitem{liu2022async}
D.~Liu \emph{et~al.}, ``Asynchronous optimisation for event-based visual
  odometry,'' in \emph{IEEE Int. Conf. Robot. Autom.}, 2022, pp. 9432--9438.

\bibitem{chamorro2022event}
W.~Chamorro, J.~Sol{\`a}, and J.~Andrade-Cetto, ``Event-based line {SLAM} in
  real-time,'' \emph{IEEE Robot. Autom. Lett.}, vol.~7, no.~3, pp. 8146--8153,
  2022.

\bibitem{kim2016real}
H.~Kim, S.~Leutenegger, and A.~J. Davison, ``Real-time 3d reconstruction and
  6-dof tracking with an event camera,'' in \emph{Eur. Conf. Comput. Vis.},
  2016, pp. 349--364.

\bibitem{hausler2019multi}
S.~Hausler, A.~Jacobson, and M.~Milford, ``Multi-process fusion: Visual place
  recognition using multiple image processing methods,'' \emph{IEEE Robot.
  Autom. Lett.}, vol.~4, no.~2, pp. 1924--1931, 2019.

\bibitem{scheerlinck2020fast}
C.~Scheerlinck \emph{et~al.}, ``Fast image reconstruction with an event
  camera,'' in \emph{IEEE Winter Conf. Appl. Comput. Vis.}, 2020, pp. 156--163.

\bibitem{scheerlinck2018continuous}
C.~Scheerlinck, N.~Barnes, and R.~Mahony, ``Continuous-time intensity
  estimation using event cameras,'' in \emph{Asian Conf. Comput. Vis.}, 2018,
  pp. 308--324.

\bibitem{lenz_gregor_2021_5079802}
G.~Lenz \emph{et~al.}, ``Tonic: event-based datasets and transformations,''
  2021, {Documentation available under \url{https://tonic.readthedocs.io}}.

\bibitem{gallego2017accurate}
G.~Gallego and D.~Scaramuzza, ``Accurate angular velocity estimation with an
  event camera,'' \emph{IEEE Robot. Autom. Lett.}, vol.~2, no.~2, pp. 632--639,
  2017.

\bibitem{gallego2018unifying}
G.~Gallego, H.~Rebecq, and D.~Scaramuzza, ``A unifying contrast maximization
  framework for event cameras, with applications to motion, depth, and optical
  flow estimation,'' in \emph{IEEE Conf. Comput. Vis. Pattern Recog.}, 2018,
  pp. 3867--3876.

\bibitem{fontan2020information}
A.~Fontan, J.~Civera, and R.~Triebel, ``Information-driven direct {RGB-D}
  odometry,'' in \emph{IEEE Conf. Comput. Vis. Pattern Recog.}, 2020, pp.
  4929--4937.

\bibitem{dauwels2008similarity}
J.~Dauwels, F.~Vialatte, T.~Weber, and A.~Cichocki, ``On similarity measures
  for spike trains,'' in \emph{Advances in Neuro-Information Processing}, 2008,
  pp. 177--185.

\bibitem{gupta2010flexible}
M.~Gupta, A.~Agrawal, A.~Veeraraghavan, and S.~G. Narasimhan, ``Flexible voxels
  for motion-aware videography,'' in \emph{Eur. Conf. Comput. Vis.}, 2010, pp.
  100--114.

\bibitem{tambe2013towards}
S.~Tambe, A.~Veeraraghavan, and A.~Agrawal, ``Towards motion-aware light field
  video for dynamic scenes,'' in \emph{IEEE Int. Conf. Comput. Vis.}, 2013, pp.
  1009--1016.

\end{thebibliography}
}

\end{document}